\documentclass[letterpaper]{article} %
\usepackage{aaai25}  %
\usepackage{times}  %
\usepackage{helvet}  %
\usepackage{courier}  %
\usepackage[hyphens]{url}  %
\usepackage{graphicx} %
\usepackage{natbib}  %
\usepackage{caption} %
\usepackage{algorithm}
\usepackage{algorithmic}

\usepackage{newfloat}
\usepackage{listings}
\DeclareCaptionStyle{ruled}{labelfont=normalfont,labelsep=colon,strut=off} %
\floatstyle{ruled}
\newfloat{listing}{tb}{lst}{}
\floatname{listing}{Listing}
\usepackage{xcolor}
\definecolor{mLightGreen}{HTML}{14B03D}

\def\ie{\emph{i.e., }}
\def\eg{\emph{e.g., }}

\newcommand{\bcolor}[2]{\color{#1}#2} %
\newcommand{\lightgreen}{\textcolor{mLightGreen}}

\newcommand{\MeasureName}{Uniform Bias}

\usepackage{amsmath}
\usepackage{amsthm}
\newtheorem{definition}{Definition}[section]
\newtheorem{theorem}{Theorem}[section]
\newtheorem{example}{Example}[section]
\newtheorem{corollary}{Corollary}[theorem]

\newtheorem*{remark}{Remark}

\usepackage{array}
\usepackage{soul}
\usepackage{subcaption} %
\usepackage{amsfonts}
\usepackage{cases}
\usepackage{algorithm}
\usepackage{algorithmic}

\usepackage{tcolorbox}		%
\tcbuselibrary{breakable,skins}		%
\tcbset{examplestyle/.style={
		enhanced jigsaw,	%
		colback=blue!08,	%
		colframe=blue!08,	%
		arc=2mm,
		boxrule=1pt,		%
		left=1mm,
		right=1mm,
		left skip=0mm,  %
		right skip=0mm, %
		top=0mm,		%
		bottom=1mm,		%
		breakable,		%
		parbox = false,		%
		before={\par\pagebreak[0]\vspace{1mm}\parindent=0pt},		
		after={\par\pagebreak[0]\vspace{1mm}\parindent=0pt},				
		bottomrule = 0mm,
		boxsep = 0mm,					%
		topsep at break=0pt,			%
		bottomsep at break=0pt,			%
		pad at break=0mm,
		pad before break=0mm,		
		pad after break=1mm,		
		bottomrule at break=0mm,
		toprule at break=0mm,		
		}}
\tcolorboxenvironment{example}{examplestyle}
\title{A Principled Approach for Data Bias Mitigation}
\author{
    Bruno Scarone\textsuperscript{\rm 1},
    Alfredo Viola\textsuperscript{\rm 2},
    Renée J. Miller\textsuperscript{\rm 3},
    Ricardo Baeza-Yates\textsuperscript{\rm 4}
}
\affiliations{
    \textsuperscript{\rm 1}Khoury College of Computer Sciences, Northeastern University, Boston, USA\\
    \textsuperscript{\rm 2}Casa de Investigadores Científicos La Comarca, La Floresta, Uruguay\\
    \textsuperscript{\rm 3}Cheriton School of Computer Science,University of Waterloo,Ontario, Canada\\
    \textsuperscript{\rm 4} Barcelona Supercomputing Center, Barcelona, Spain\\
    scarone.b@northeastern.edu, alfredo.viola@gmail.com, rjmiller@uwaterloo.ca, rbaeza@acm.org
}

\begin{document}

\maketitle

\begin{abstract}
The widespread use of machine learning and data-driven algorithms for decision making has been steadily increasing over many years. \emph{Bias} in the data can adversely affect this decision-making. 
We present a new mitigation strategy to address data bias. Our methods are explainable and come with mathematical guarantees of correctness. They can take advantage of new work on table discovery to find new tuples that can be added to a dataset 
to create real datasets that are unbiased or less biased. Our framework covers data with non-binary labels and with multiple sensitive attributes. Hence, we are able to measure and mitigate bias that does not appear over a single attribute (or feature), but only intersectionally, when considering a combination of attributes. We evaluate our techniques on publicly available datasets and provide a theoretical analysis of our results, highlighting novel insights into data bias.

\end{abstract}

\section{Introduction}\label{sec:intro}

In this work, we take advantage of new advances in table discovery (specifically, table discovery in data lakes~\cite{DBLP:journals/pvldb/NargesianZMPA19,DBLP:journals/vldb/ChapmanSKKIKG20}) to consider the problem of bias mitigation. Given a biased dataset, can we modify it to be unbiased or less biased? We do this in the context of real data where we do not want to simply modify the data to make it unbiased (for example, by changing the values associated with a tuple in a protected group). Our mitigation strategies use real data rather than synthetically altering data. We also do this in a context where we want to use the data for real data analysis. Hence, just removing tuples until we get a less biased subset that matches our fairness goal may not yield sufficient data to perform an analysis (such as training a machine learning (ML) model). In some cases, we may need to use table discovery to find new (real) data to meet our goals.  An important contribution of our work is to help a data scientist explore the space of possible mitigation solutions that make the data less biased.

There are two main purposes when measuring data bias. On the one hand, we may be interested in quantifying the bias of a dataset as an assessment of its quality, given that bias provides information about the data's representativeness\footnote{This can be referred to as an ideal scenario of reference.} or completeness (a well-studied data quality dimension \cite{Batini2016-cr}). But arguably, the fundamental practical application of such a metric is, upon detection of a significant bias, to obtain an unbiased  (or less biased) dataset. We refer to algorithms that use data discovery to add tuples and/or use other table transformations (including tuple deletion or modification) to obtain a less biased dataset as bias mitigation algorithms. An underlying principle in bias mitigation is to perform a minimal change to a dataset that is sufficient to achieve a fairness goal.  

\subsection{Motivating Example}\label{sec:intro_example}

A major challenge in data fairness is to ensure that the dataset used for analysis has an appropriate representation of relevant demographic groups~\cite{Nargesian2021}. This is because insufficiently representative training data has been repeatedly shown to be extremely problematic in a wide range of ML application domains \cite{hort2024biasmitigationsurvey,pagano2023bias}.

Consider a US Bank that decides to build an ML model for default loan prediction, i.e., predicting the probability that a person pays back a given loan. Since the bank does not have enough internal quality data, they decide to use the Adult Dataset \cite{misc_adult_2} for this purpose, a widely used dataset containing demographic information (14 attributes) from several thousand individuals, including an attribute  indicating if a person's annual income exceeds $\$50,000$ or not. The bank decides to grant loans to people whose annual income exceeds $\$50,000$. We will call these tuples positively labeled or positive. In what follows, we consider the binary gender attribute (with values Male or Female) of this dataset.  Tuples with the value female form a {\em protected} group and we call them protected tuples.

Upon inspection, the analysts realize that for the analysis they want to perform (predicting likelihood of paying back a loan), the data is biased against women: there are $1,179$ out of the $10,771$ tuples representing women that have a yearly income greater than $\$50,000$ ($10.9\%$), while this fraction is $6,662/21,790$ for men ($30.6\%$). The data scientists are worried that if they use this version of the data, the resulting model may grant fewer loans to women in a discriminatory way (in cases where they could indeed have paid the loan back), so they aim at constructing a new version of the dataset that is group fair w.r.t. gender.

Since the total number of tuples is $n = 32,561$ and there are $p= 10,771$ women,  we can modify the dataset so that $3,296$ of the women are positive (and hence at parity with the men) or we can lower the number of positive men to $2,375$, (so men are at parity with the women). 
Preprocessing techniques like feature normalization 
are common practice in ML. However, it is not acceptable for data scientists and domain experts to arbitrarily change attribute values from the tuples of the dataset. In our running example, it does not make sense to change the address, race or gender of a person. There are also cases where it is unreasonable to significantly change the salary value of an individual leaving the loan decision unchanged.
Thus in many analyses, the only realistic operations are tuple additions (with new real data) and deletions. This turns even the fairness estimation task into a more challenging task as parameters like the number of tuples are no longer constant. Returning to our example, an additional requirement of the ML engineers is to have $n\geq 30,000$ for the learned model to have reasonable performance, hence only deleting tuples is not an option.

\begin{figure}[t]
    \begin{subfigure}{0.45\linewidth}
      \includegraphics[scale=.4]{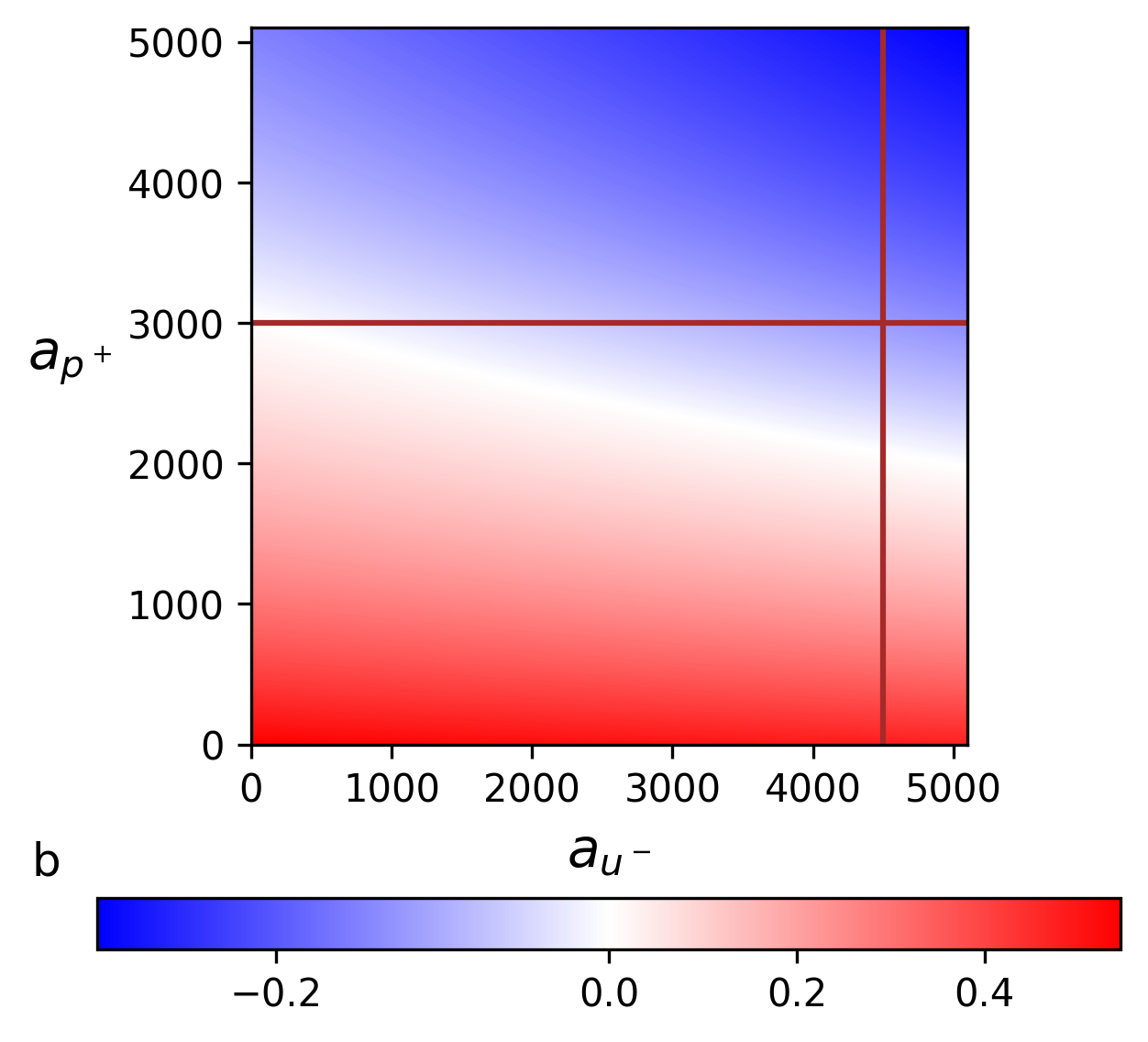}
      \caption{\small Adding positive protected \newline \hspace*{0.4cm} and unprotected tuples.\newline ~}
    \label{fig:adult_2varspols_sub1}
    \end{subfigure}
    \hspace*{.1cm}
    \begin{subfigure}{0.45\linewidth}
    \includegraphics[scale=.4]{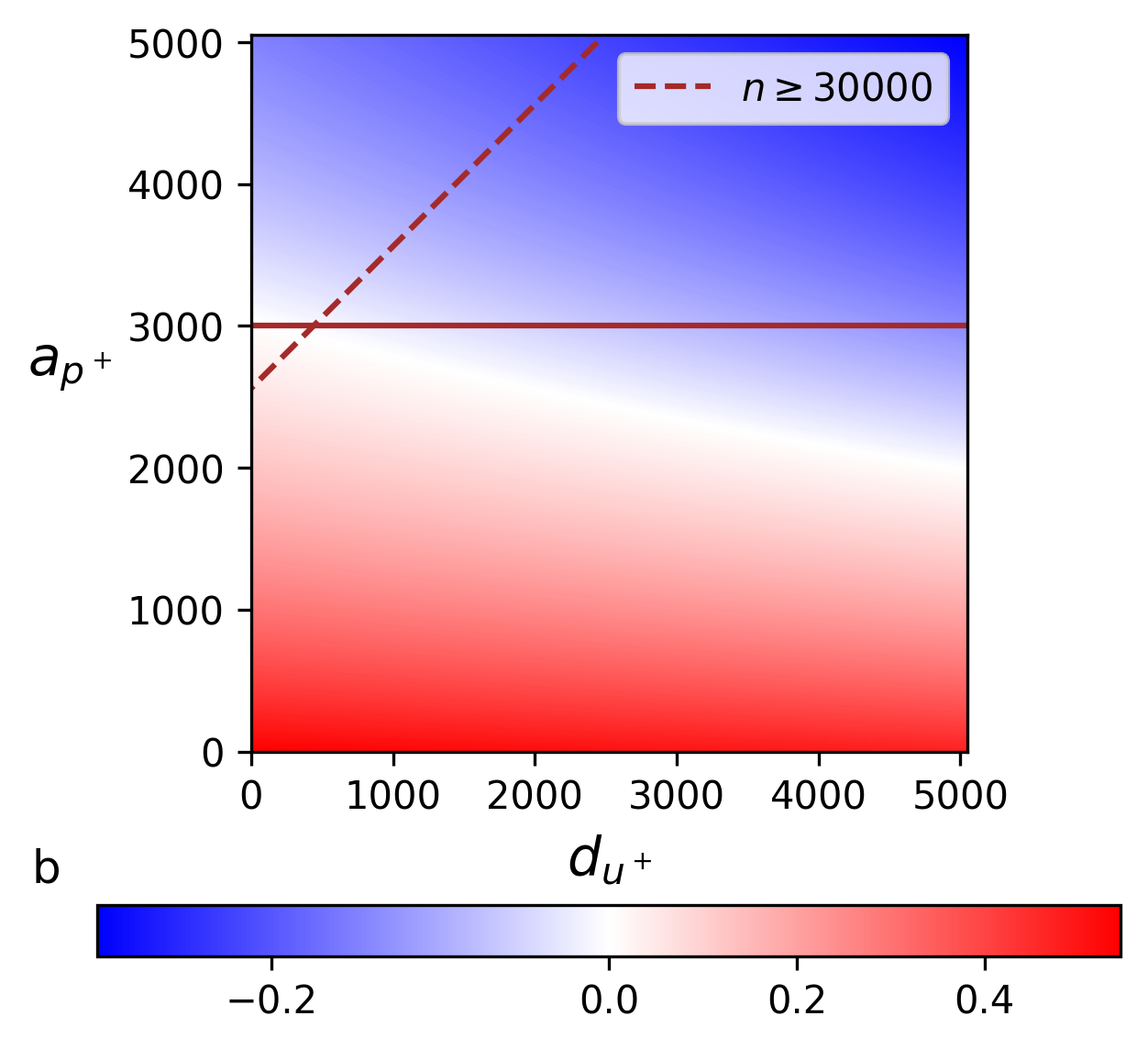} 
    \caption{\small Adding positive protected \newline \hspace*{0.4cm} and deleting negative \newline \hspace*{0.6cm} unprotected tuples.}
    \label{fig:adult_2varspols_sub2}
    \end{subfigure}
    \caption{Adult Dataset - Policies using two operations.}
    \label{fig:adult_2varspols}
    \vspace*{-0.5cm}
\end{figure}

After analyzing the available data sources (open source data lakes as well as data available through data brokers) the team determines that they can find at most $3,000$ positive protected tuples and $4,500$ negative unprotected tuples. Using our approach, the data scientists are able to simulate different scenarios to determine, given the number of added and deleted tuples of both types, how far different mitigation strategies (different tuple additions and deletions) end up from creating a fair dataset. The results are shown 
in Figure \ref{fig:adult_2varspols}, where $b=0$ indicates that the resulting dataset is group fair (white region in Figure \ref{fig:adult_2varspols}).

In Figure \ref{fig:adult_2varspols_sub1}, we model adding positive protected tuples ($a_p^+$, depicted on y-axis) and adding negative unprotected tuples ($a_u^-$, depicted on x-axis); while on Figure \ref{fig:adult_2varspols_sub2} we instead consider deleting positive unprotected $d_u^+$ for the second variable. In these figures, white indicates group fairness. Red indicates a bias against women (as in the original dataset) and blue is a bias against men. Here, the horizontal line $a_p^+=3,000$ represents the constraint that there are at most $3,000$ new positive protected tuples that can be added. The vertical line $a_u^-=4,500$ represents the same for negative unprotected tuples. 
The diagonal line in Figure \ref{fig:adult_2varspols_sub2} represents the restriction of needing at least $30,000$ tuples in the resulting dataset. Thus, in Figure \ref{fig:adult_2varspols_sub1}, the feasible region (datasets we can construct under the problem's restrictions) is 
the large lower left quadrant that includes the origin, while for Figure \ref{fig:adult_2varspols_sub2} it is the small triangular shape including the point $(0,3000)$. Interestingly, Figure \ref{fig:adult_2varspols_sub1} shows how adding negative unprotected tuples ($a_u^-$, men who earn less than 50K) reduces the number of positive protected tuples required ($a_{p^+}$), but in a non-obvious way. For example, by adding $4,500$ negative unprotected tuples, we reduce the number of protected positive tuples needed to $2,077$.  Figure \ref{fig:adult_2varspols_sub2} shows that by deleting a few of the positive unprotected tuples, we can add fewer than the available $3,000$ new positive protected tuples, but the space of options is much more limited. In this paper, we introduce mathematically sound methods to perform this process in an efficient and interpretable way.\footnote{The analytical solution of this example is included in the Appendix (Section \ref{sec:costs}).} 

\subsection{Contributions}%

The contributions of our work are summarized as follows:  
\begin{itemize}
    \item We introduce \MeasureName\ (UB), the first 
    intersectional, multi-label bias measure that can be computed directly from a given dataset $T$. %
    Our measure is interpretable and can model bias in datasets with multiple classes (multi-label) and with multiple protected attributes.
    \item In contrast to the hundreds of papers surveyed by Hort et al.~\cite{hort2024biasmitigationsurvey}, we formally define bias for multi-class problems with multiple non-binary protected attributes.
    \item We present a new mitigation strategy to address data bias. Our strategies are interpretable and explainable. They can take advantage of new work on table discovery to create datasets that meet a fairness goal. Our mitigation algorithm guarantees a label frequency preservation property for the protected groups, meaning that the ratios of tuples with a given label (\eg people who are given a loan) is the same after bias mitigation is applied, which we argue is relevant for practical applications.
    \item We show how UB solves existing issues of anti employment discrimination rules used by the Office of Federal Contract Compliance Programs.
    \item 
    We evaluate our techniques on real datasets (recommended by a recent comprehensive survey~\cite{hort2024biasmitigationsurvey}) and show that our mitigation strategies can be used in practice to produce unbiased training data that do not significantly lower the accuracy of ML models, and in some cases improves their performance. %
\end{itemize}

The rest of this paper is organized as follows. In Section~\ref{sec:rel_work} we detail related work. In Section~\ref{sec:prelim} we present the notation. In Section~\ref{sec:defining_bU} we introduce \MeasureName\, our new bias measure. Section \ref{sec:motivation} motivates the significance of \MeasureName{} and how it can be used to solve open problems identified in the literature. In Section~\ref{sec:intersect} we extend our approach to be intersectional and to consider non-binary labels. Section \ref{sec:mitigation} uses these ideas for bias mitigation in this new context. Section \ref{sec:extensions} extends our techniques in ways that are relevant in practice, while in Section \ref{sec:mleval} we evaluate how our mitigation affects the performance of ML models. We end with our conclusions and future work in Section \ref{sec:conclusion}.

\section{Related Work}
\label{sec:rel_work}

Nowadays, the concept of \emph{bias} is prevalent in computer science literature. However, although the seminal paper on this topic is from 1996 \cite{bias1996}, ten years ago was not the case. Bias is used in heterogeneous settings, where the underlying meaning depends on the context. In some of these contexts, such as statistics, the definition is rigorous, while in others (\eg data mining, machine learning, web search and recommender systems) it is either handled informally or without a consensus within the broader community. This is in part due to the complexity and interdisciplinary nature of the problem, as noted by Zliobaitè \cite{vzliobaite2017measuring}. The common idea across all these notions is that of a \emph{systematic deviation from a predefined reference value}. 
A closely related concept is that of fairness, defined as the absence of (negative) discrimination.  %

\subsection{Measuring Bias} 

In the context of data mining and ML, where one of the main goals is to design fair models for the task at hand, researchers have proposed a variety of heuristic measures~\cite{mehrabi2021survey}
with the objective of quantifying bias, and thus being able to design new algorithms that would optimize these measures \cite{vzliobaite2017measuring}. These measures can be classified into individual and group/statistical measures.\footnote{Sometimes the definition only specifies the fairness/unbiased condition without quantifying the deviation w.r.t. this value.} Individual fairness (introduced by Dwork et al. \cite{dwork_fairness_through_aw}) centers on the idea that similar individuals should be treated similarly.\footnote{This includes Counterfactual or Causal based fairness \cite{salimi2020database,plecko2022causal}.} On the other hand, group fairness focuses on treating different demographic groups equally. The heterogeneous unprincipled nature of the measures and evaluation approaches makes it difficult to compare results (see Osoba et al.~\cite{linkedin} and Lum et al.~\cite{lum2022debiasing} as recent examples), as well as to establish guidelines for practitioners and policymakers. Verma et al. \cite{Verma2018} collects several individual and group  measures and computes each of them on a case-study intuitively explaining why the same case can be considered fair according to some definitions and unfair
according to others. Yeh et al. \cite{10.1145/3630106.3658922} focuses on two common families of statistical measures (ratio and difference based), theoretically analyzing their relationship and providing empirical results to establish initial guidelines for which can be used in different contexts. With the goal of providing a unifying view, Zliobaitè \cite{vzliobaite2017measuring} surveys and categorizes various statistical measures, experimentally analyzes them, and recommends which ones to use in different contexts. Its foundational notions consist of precisely defining the condition for a dataset to be \emph{unbiased}, as well as a (randomized) bias addition algorithm. 
It is important to note that in this context, statistical measures are used as a tool to indirectly quantify the effect of added bias on tables.  

\subsection{Intersectionality}

Crenshaw introduced the term intersectionality \cite{crenshaw2013demarginalizing}, highlighting that the discrimination experienced by Black women is greater than the sum of racism and sexism individually and thus establishing that any bias analysis that does not take the intersection of sensitive groups into account cannot sufficiently address the particular manner in which individuals experience discrimination. In the context of data fairness, this translates into the fact that it is not enough to verify fairness for sensitive attributes independently. In fact, we can construct datasets that are group fair w.r.t. to gender and race, but not w.r.t. their intersections.
This was also shown in the context of fair classification by Kearns et al. \cite{kearns2019anempiricalstudy}. Wang et al. \cite{wang2022towards} remark that research in fair ML has historically considered a single binary demographic attribute, and they study how to address intersectionality from a practical perspective in a ML pipeline.

\subsection{Bias Mitigation Methods}
In a recent survey, Hort et al. \cite{hort2024biasmitigationsurvey} provide a comprehensive analysis of 341 publications concerning bias mitigation methods for ML classifiers. The authors identified three types of bias mitigation methods: Pre-processing, where bias mitigation is applied to the training data to prevent it from reaching ML models; In-processing, where bias mitigation is done while training the models and Post-processing, where it is done on trained models. Our method falls into the pre-processing category. The authors remark that there are almost twice as many publications with in-processing methods than pre-processing and as Salimi, Howe, and Suciu~\cite{salimi2020database} highlight, one of the advantages of these methods is that they can be used in conjunction with any ML model.

Hort et al.~remark that consolidating a common set of metrics is still an open challenge. Another related problem they highlight is to make sure that the metrics used are representative for the problem at hand and they state that future work should focus on multi-class problems, and non-binary sensitive attributes, which was mentioned by only 15 out of the 341 publications they considered. Out of these 15 publications, only one \cite{alabdulmohsin2022reduction} can deal with both non-binary sensitive attributes and multi-class predictions, which is what we do with our UB (Uniform Bias) measure and our mitigation algorithm.

Another challenge identified by Hort et al.~is to include trade-offs when dealing with accuracy and/or multiple fairness metrics. We evaluate our model on real datasets and show that our mitigation strategies can be used to produce training data that yield better performing ML models. In particular, we use the first two and seventh most widely used datasets according to Table 9 in \cite{hort2024biasmitigationsurvey}.

Providing fairness guarantees is considered a relevant avenue of future work, stating that allowing for interpretable and explainable methods can aid in this regard. We provide mathematical guarantees in the most general case for both the interpretation of UB and our mitigation algorithm. Additionally, our techniques allow a user to specify acceptable or allowed levels of bias (\eg determined by domain experts) in a natural way, which is also highlighted as one of the challenges by Hort et al. The survey also notes the importance of having continuous implementations of bias mitigation methods in real-world scenarios. Our methods are simple and can be naturally implemented to continuously monitor the bias level of data and compute mitigated versions.

\subsection{Our Approach} 
We present a new intersectional data bias measure that can directly be computed from the target table and argue why it improves the state-of-the-art. 
Namely, UB is the first measure that can handle an arbitrary number of multi-valued sensitive attributes and a multi-valued label in a simple way. The same ideas used to derive UB can serve as the basis for explainable and mathematically guaranteed bias mitigation strategies that can be computed for any dataset. We show how our bias mitigation produces new versions of well known datasets that can be used to improve the performance of ML models.

\section{Preliminaries}
\label{sec:prelim}

As is common, we denote variables (\ie, dataset attributes) by uppercase letters, $X$, $Y$, $Z$; their values with lowercase letters, $x$, $y$, $z$; and denote vectors (respectively, sets) of variables (values) using boldface ($\textbf{X}$ or $\textbf{x}$). The domain of variable $X$ is $Dom(X)$, the domain of a vector of variables is $Dom(\textbf{X})=\prod_{X\in \textbf{X}} Dom(X)$.  
To simplify notation in formulas, when we use a value \textbf{x} we mean $|\textbf{x}|$, which is the number of tuples from the dataset with value \textbf{x} and we omit the attribute \textbf{X} when it is clear from context. 
\subsection{Bias Notation}
We start by considering the classical setting of algorithmic fairness where we have a classification task (\eg hiring men and women candidates), where tuples have a target (class) attribute $Y$ with a binary domain ($y\in\{0,1\}$).  In the literature, the values of $Y$ are often called \emph{binary target labels.} In addition, tuples have a binary \emph{sensitive or protected attribute} $S$ defining a protected  (or unprivileged) group $s=1$ and an unprotected  (privileged) group $s=0$.  
This simplified scenario has been extensively studied in the fair data mining and machine learning literature~\cite{vzliobaite2017measuring, wang2022against, hort2024biasmitigationsurvey}. For this reason, Table \ref{tab:notation} includes the the notation for this binary scenario. %

However, our work naturally extends to  multi-class problems ($dom(Y)$ can be any set of values) and to account for multiple sensitive attributes ($\textbf{S}$) with non-binary values ($dom(\textbf{S})$ can be any set of values). Let $|\textbf{S}| = m$, then $\textbf{s} \in (dom(S_1)\cup \{\epsilon\})\times ... \times (dom(S_m)\cup \{\epsilon\})$ denotes a possible protected group. Here, $\epsilon$ represents all values in a given domain. For example, for attributes gender = \{male, female, non-binary\}, and age = \{young, middle-aged, retired\}, one possible protected group (non-binary, retired) represents all tuples that are both non-binary and retired, while another possible protected group (male, $\epsilon$) represents all men independent of age. We drop the parenthesis and the $\epsilon$ when the context is clear, \ie use ``male" instead of (male, $\epsilon$) to represent the group of males.
Table~\ref{tab:notation} also defines notation for this more general multi-class, non-binary case.
All quantities refer to a given dataset $T$. Note that our measure, given a sensitive group $s$ (or $\textbf{s}$ in the non-binary case) and a class label $y$, can be computed only based on the table $T$, that is, our bias measure $\text{UB}(T,\textbf{s},y)$ as defined in Section~\ref{sec:defining_bU}. This is also the case for other preexisting measures we will reference later and that are included in Table \ref{tab:notation} (IR and OR).%

\begin{table}[t]
    \centering
    \begin{tabular}{|c|>{\centering\arraybackslash}p{6.5cm}|}
        \hline
        \textbf{Variable} & \textbf{Definition} \\
        \hline
        $n$ & Total number (\#) of tuples ($|T|$) \\
        \hline
        $n^+$  & For binary problems, \# positive tuples, $\{t\in T:t[Y]=1\}$\\
        \hline
        $n^y$ & For multi-class problems, \# tuples with a given class label $y$: $\{t\in T:t[Y]=y\}$\\
        \hline
        $p$ & For binary problems, \# tuples in the protected group, $\{t\in T:t[S]=1\}$ (resp., for  unprotected group, $u$)\\
        \hline
        $p^+$ & For binary problems, \#  positive protected tuples  $\{t\in T:t[Y]=1 \wedge t[S]=1\}$(resp., $u^+$)\\
        \hline
        $P$ & Protected ratio, $p/n$ (resp., $U=u/n$) \\
        \hline
        $f_{p,+}$ & $p^+/p$\\
        \hline
        $f_y$ & $n^y/n$\\
        \hline
        $f_{\textbf{s},y}$ & Ratio of \# tuples with protected attributes \textbf{s} and class label $y$ over all tuples in the protected group \textbf{s}:
        $|\textbf{s}y|/|\textbf{s}|$\\
        \hline
        $\lfloor i \rceil$ & Closest integer to $i$, $\lfloor i+0.5\rfloor$\\
        \hline
        IR & $\frac{f_{p,+}}{f_{u,+}}=\frac{p^+\cdot (n-p)}{p\cdot (y^+-p^+)}$\\ %
        \hline
        OR & $\frac{1-f_{p,+}}{f_{p,+}}\cdot\frac{f_{u,+}}{1-f_{u,+}}=\frac{u^+/u^-}{p^+/p^-}$\\
        \hline
        MD & $f_{u,+}-f_{p,+}$ \\
        \hline
    \end{tabular}
    \caption{Notation used in this paper.}
    \label{tab:notation}
    \vspace*{-.3cm}
\end{table}

\begin{table*}[t]
    \centering
    \small
    \begin{tabular}{|l|c|c|c|c|c|c|c|c|c|} %
        \hline
        Dataset & $n$ & $n^+$ & $p$ & $p^+$ & $p^+(0)$ & UB & IR & OR & MD\\
        \hline
        Adult \cite{misc_adult_2}& 32561 & 7841 & 10771 & 1179 & 2594 & 0.55 & 0.36 & 3.58 & 0.2\\
        \hline
    \end{tabular}
    \caption{Summary statistics of the Adult dataset.}
    \label{tab:datasets}
\end{table*}

\section{\MeasureName}
\label{sec:defining_bU}

In this section, we present an example to motivate and introduce our new bias measure. Consider the setting where a company hires $n^+=200$ new employees from a set of $n=600$ people, where $p=150$ applicants are women (protected) and $u=450$ are men (unprotected), thus gender is the protected attribute ($S$).

\begin{table}[t]
    \centering
    \begin{tabular}{|c|c|c||c|c||c|c|c|}
        \hline
          & $p^+$ & $u^+$ & $f_{p,+}$ & $f_{u,+}$ & $UB$  & $IR$ & $MD$\\
        \hline
        $T_0$ & $50$ & $150$ & $1/3$ & $1/3$ & $0$ & $1$ & $0$\\
        \hline
        $T_1$ & $40$ & $160$ & $4/15$ & $16/45$ & $1/5$ & $3/4$ & $4/45$\\
        \hline
        $T_2$ & $60$ & $110$ & $2/5$ & $11/45$ & $-1/5$ & $18/11$ & $-7/45$\\
        \hline
    \end{tabular}
    \caption{Summary statistics and measures for three classes of tables. For all rows we have: $n=600$, $y^+=200$, $p=150$, $p^+(0)=50$, $u^+(0)=150$ and $f_+=1/3$.}
    \label{tab:mot_example}
    \vspace*{-0.3cm}
\end{table}

Suppose that after this selection process concludes, we have access to the corresponding tabular data showing the distribution of accepted and rejected candidates. There are three possible scenarios of interest, whose summary statistics are shown in Table \ref{tab:mot_example}:\footnote{Note that if we only consider the proportions ($f$, $f_p$ and $f_u$), then we can associate each row in Table \ref{tab:mot_example} to the set of tables whose parameters satisfy the proportions, which makes the analysis much more general (\eg independent of $n$).}

\begin{itemize} 
    \item $T_0$: The proportion of hired women ($f_{p,+}$) and men ($f_{u,+}$) are equal and coincide with the fraction of positive tuples of the population ($f_+$). Here, $f_{p,+} = f_{u,+} = f_+$, and we say $T_0$ is \textbf{unbiased}, following what is common practice in the algorithmic fairness community \cite{mehrabi2021survey};
    \item $T_1$: The proportion of hired women ($f_{p,+}$) is \textbf{lower} than that of men ($f_{u,+}$). Here, $f_{p,+} < f_+ < f_{u,+}  $, thus $T_1$ exhibits a \textbf{negative bias} against women (protected group).
    \item $T_2$: The proportion of hired women ($f_{p,+}$) is \textbf{higher} than that of men ($f_{u,+}$). Here, $f_{u,+} < f_+ < f_{p,+} $, and so $T_2$ exhibits a bias against men (unprotected group) or equivalently a \textbf{positive bias} in favor of women.
\end{itemize}

Now we introduce ideas to quantify bias from the viewpoint of a specific protected group, using the notation defined in Table \ref{tab:notation}. Consider the case of $T_1$. Table $T_0$ gives the ideal number of hires, which we call $p^+(0) = 50$. Ideally we would have wanted for $p^+(0)=50$ women to get hired, but only $p^+=40$ were. Our goal is to quantify this bias, which is clearly related to the difference or deviation of $10$ from the unbiased state $T_0$. We start by observing that this difference is $20\%$ of the target quantity $p^+(0)=50$, \ie $20\%$ less women are being hired than desired. Thus, if we take $b$ (the bias) to be this percentage we get $p^+=p^+(0)-b\cdot p^+(0)$ with $b=0.2$. 
If $b=0$ we have unbiased data and if $b=1$ negative bias against women is maximized. As stated before, representing these ideas in terms of ratios will be useful. Thus, we divide both sides of the last equation by $p$ and noting that 
when $b=0$, then $f_{p,+}=f_+$ (unbiased condition), we derive the following expression $f_{p,+}=(1-b)\cdot f_{+}$. Solving for $b$, we have a measure of the percentage of missing elements ($20\%$ for $T_1$). These ideas are the building blocks for our proposed formal bias definition.

\begin{definition}[\MeasureName \ measure, one binary attribute, binary label]\label{def:bU}
    Given any table $T$, its \emph{\MeasureName} (UB) w.r.t. the group $s\in \{p,u\}$ and label $y\in\{+,-\}$ is given by
    \[ \text{UB}(T,s,y) = b_{s,y} =  1-\frac{f_{s,y}}{f_y}~.
    \]
    Observe that the right side can be directly computed based on the data and that it is linear w.r.t. $f_{p,+}$ and if $f_{p,+}$ is proportionally greater than (or less than) $f_+$ by the same about, the bias will be the same but negated (for example, 20\% or -20\%).  
    Additionally, note that by writing the definition in terms of relative table quantities, all tables with the same values of $f_{p,+}$ and $f_+$ will result in the same UB, regardless of their size. For brevity, we use the symbol $b$ when the context ($T,s,y$) is clear.
\end{definition}

Note that, as discussed before, when $b_{p,+}=0$ we have $f_{p,+}=f_+$, when $b_{p,+}>0$ we have $f_{p,+}<f_+$ and when $b_{p,+}<0$ we have $f_{p,+}>f_+$. 

We say UB is uniform because it is not context dependent and the measure will be the same for any dataset $T$ with equal relative parameters ($f_{p,+}$ and $f_+$). 
Finally, as remarked in the literature \cite{Verma2018}, in practice one does not expect a fair dataset to have a bias that is exactly zero. In this sense, one can determine an admissible range of bias values determined by domain experts (\eg $|b|\leq .1$) to declare a table to be sufficiently unbiased or fair.

\section{Comparing UB to existing measures}
\label{sec:motivation}

The work by Gastwirth \cite{gastwirth2021summary} centers around rules issued by the Office of Federal Contract Compliance Programs\footnote{This office oversees employment practices and promotes efforts to diversify the work forces of government contractors in the USA.} (OFCCP) in November 2020 to resolve employment discrimination issues. In this context, Gastwirth presents an in-depth analysis of how the agency will use and evaluate statistical evidence in its monitoring of government contractors’ compliance with equal employment laws. 
The rules state that the agency will ordinarily use the impact ratio as its measure of practical significance and uses what is known as the ``fourth-fifths rule'' (used since 1970) to detect violations. As explained by Oswald et al. \cite{oswald2016measuring}, the four-fifths rule is violated when the selection rate of one applicant group (\eg Hispanic) is less than 80\% of the selection rate for the group with the highest rate (\eg White). 

Gastwirth observes that while the rules develop the agency’s classification system in terms of the impact ratio, they also
allow it to use other measures such as the odds ratio. In terms of Practical Significance measures\footnote{Practical significance is given more attention in the final rules used by the OFCCP than in the original proposal \cite{gastwirth2021summary}.}  (including both the impact and odds ratio), there is an extensive literature (\eg \cite{gastwirth2021summary,oswald2016measuring} and the references therein) analyzing these measures and illustrating their flaws in the context of determining disparate impact. However, these works do not present a systematic approach to study them (\eg to precisely characterize when they do not work and why) nor to solve the identified problems.

In order to show the limitation of IR and the mean difference (MD, Table \ref{tab:notation}) in not being able to distinguish datasets that are significantly different in terms of disparate impact, we construct two summary statistics (Tables \ref{tab:SS_IR_MD_1} and \ref{tab:SS_IR_MD_2}) that have a fix IR and MD, but for which the values of UB (denoted by $b$) vary significantly. We take IR and MD as representatives of ratio based and difference based measures.

\begin{table}[t!]
    \centering
    \begin{tabular}{|c|c|c|c|c|c|c|}
        \hline
        $p^+$ & $p$ & $u^+$ & $u$ & $f_+$ & $p^+(0)$ & $b$\\
        \hline
        $396$ & $990$ & $5$ & $10$ & $.401$ & $396.99$ & $\lightgreen{.0025}$\\
        \hline
        $388$ & 970 & 15 & 30 & .403 & 390.91 & \lightgreen{.0074} \\
        \hline
        $360$ & 900 & 50 & 100 & .410 & 369.00 & \lightgreen{.0244} \\
        \hline
        $320$ & 800 & 100 & 200 & .420 &  336.00 & \lightgreen{.0476} \\
        \hline
        $232$ & 520 & 210 & 420 & .442 &  229.84 & \lightgreen{.0950} \\
        \hline
        $160$ & 400 & 300 & 600 & .460 &  184.00 & \bcolor{red}{.1304} \\
        \hline
        $40$ & 100 & 450 & 900 & .490 &   49.00 & \bcolor{red}{.1836} \\
        \hline
        $12$ & 30 & 485 & 970 & .497 & 14.91 & \bcolor{red}{.1951} \\
        \hline
        $4$ & 10 & 495 & 990 & .499 & 04.99 & \bcolor{red}{.1983} \\
        \hline
    \end{tabular}
    \caption{\footnotesize Summary statistics showing constant $IR$ and $MD$, while $b\in [0.25\%,19.83\%]$. For all rows we have: $n=1000$, $f_p=.4$, $f_u=.5$, $IR=.8$ and $MD=.1$.}
    \label{tab:SS_IR_MD_1}
\end{table}

\begin{table}[t!]
    \centering
    \begin{tabular}{|c|c|c|c|c|c|c|}
        \hline
        $p^+$ & $p$ & $u^+$ & $u$ & $f_+$ &  $p^+(0)$ & $b$\\
        \hline
        199  & 995 & 4 & 5 & .203 &  201.985 & \lightgreen{.0148} \\
        \hline
        194 & 970 & 24 & 30 & .218 &  211.46 & \lightgreen{.0826} \\
        \hline
        180 & 900 & 80 & 100 & .260 & 234.0 & \bcolor{red}{.2308} \\
        \hline
        100 & 500 & 400 & 500 & .500 & 250.0 & \bcolor{red}{.6000} \\
        \hline
        40 & 200 & 640 & 800 & .680 & 136.0 & \bcolor{red}{.7059} \\
        \hline
        20 & 100 & 720 & 920 & .740 &  74.0 & \bcolor{red}{.7297} \\
        \hline
        10 & 50 & 760 & 950 & .770 & 38.5 & \bcolor{red}{.7403} \\
        \hline
        1 & 5 & 796 & 995 & .797 & 3.9850 & \bcolor{red}{.7491} \\
        \hline
    \end{tabular}
    \caption{\footnotesize Summary statistics showing constant $IR$ and $MD$, while $b\in [1.48\%,74.91\%]$. For all rows we have: $n=1000$, $f_p=.2$, $f_u=.8$, $IR=.25$ and $MD=.6$.}
    \label{tab:SS_IR_MD_2}
    \vspace*{-0.5cm}
\end{table}

In these examples we assume that an absolute value of UB lower than 10\% denotes an acceptable disparity rate.
For Table \ref{tab:SS_IR_MD_1}, IR (0.8) indicates that disparate impact is unlikely (per the four-fifths rule, \cite{oswald2016measuring,gastwirth2021summary}). Meanwhile, the IR (0.25) shown in Table \ref{tab:SS_IR_MD_2} is considered to strongly suggest this kind of discrimination (as presented in \cite{oswald2016measuring,gastwirth2021summary}). %
Nevertheless, in both tables there are cases that, according to UB, either present strong discrimination or barely any discrimination at all. In our view, it is essential to provide a rigorous explanation of this.

For constructing these tables we start by fixing $f_{p,+}$ and $f_{u,+}$, which in turn determine the values of IR and MD. For a fix $n$, by varying $p$, the quantities $p^+$, $u^+$ and $f$ are determined for each row. %
Notice that when $p$ is large (resp. for $u$), most of the table is filled according to $f_{p,+}$ (resp. for $f_{u,+}$). As a consequence, when $p$ is large, $p^+$ is close to the optimum $p^+(0)$ and so $b$ is small. On the other hand, when $p$ decreases ($u$ increases), since $f_{u,+}>f_{p,+}$, $p^+$ gets further away from $p^+(0)$ and thus $b$ increases. Recall we define the protected ratio $P = p/n$. Given that $IR(b)=\frac{1-b}{1+b\cdot P/U}$ (Table \ref{tab:notation}), it is easy to see that when $P\rightarrow 1$, $b\rightarrow 0$, and when $P\rightarrow 0$ then $b\rightarrow 1-IR$.

The key observation is that $IR$ and $MD$ are not sufficiently descriptive, since they do not distinguish what is happening in each row. In fact, the $b$ values in these rows present completely different scenarios in terms of the disparate impact they describe. Precisely, in Table \ref{tab:SS_IR_MD_1}, $b$ varies in the range $[0.25\%, 19.83\%]$, while in Table \ref{tab:SS_IR_MD_2} this range becomes $[1.48\%,74.91\%]$. Given the interpretation of $b$ as $p^+$ being a $1-b$ fraction from $p^+(0)$, it is crucial to develop new disparate impact measures that are able to detect these significant disparities across the rows. Our measure UB achieves this goal.   

Since $f_{p,+}$ and $f_{u,+}$ are constant, this variation is due to $f_+$ that ranges from $f_{p,+}$ when $P\rightarrow 1$ to $f_{u,+}$ when $P\rightarrow 0$. Neither IR nor MD are sensitive to $f_+$, while UB is. While these flaws in IR and MD (as well as for other measures) have been extensively pointed out in the literature, to the best of our knowledge, this is the first time a systematic explanation of the flaws has been presented.
Based on this explanation, we believe UB to be a more suitable measure to use for determining disparate impact when analyzing data.

To further illustrate the practical impact of UB we present now a solution to the contradictory judgments identified by Oswald {\em et al.} in Table 5.3 of \cite{oswald2016measuring}, arising from the use of IR, OR and MD. %
For instance, when looking at the case in Table \ref{tab:oswald_IR_OR_b} where $MD=.2$ and $f_+=.1$, although neither IR nor OR indicate an adverse (bias) impacting the data, UB (18\%) does. Furthermore, when $MD=.10$ and $f_+=.5$, IR and OR even disagree on their judgments. Meanwhile, UB provides additional evidence supporting the claim made using OR. Moreover, the key observation is that $b_{p,+}=18\%$ can be interpreted as the fraction of tuples away from the optimum $p^+(0)$. As far as we are aware, none of the measures presented in the literature give a similar quantitative explainable interpretation. This example provides further evidence that UB is a good metric to be used in this context. 

\begin{table}[t]
    \centering
    \begin{tabular}{|c|c|c|c|c|c|}
        \hline
         \multicolumn{3}{|c|}{$MD = .02$} & $IR$ & $OR$ & $b_{p,+}$\\
        \hline
        $f_+=.5$ & $f_{p,+}=.482$ & $f_{u,+}=.502$ & \lightgreen{.960} & \lightgreen{.923} & \lightgreen{.036}\\
        \hline
        $f_+=.1$ & $f_{p,+}=.082$ & $f_{u,+}=.102$ & \lightgreen{$.804$} & \lightgreen{$.786$} & \bcolor{red}{$.18$}\\
        \hline
        \hline
        \multicolumn{3}{|c|}{$MD = .05$} & $IR$ & $OR$ & $b$\\
        \hline
        $f_+=.5$ & $f_{p,+}=.455$ & $f_{u,+}=.505$ & \lightgreen{$.901$} & \lightgreen{$.818$} & \lightgreen{$.09$}\\ 
        \hline
        $f_+=.5$ & $f_{p,+}=.055$ & $f_{u,+}=105$ & \bcolor{red}{$.542$} & \bcolor{red}{$.496$} & \bcolor{red}{$.45$}\\
        \hline
        \hline
        \multicolumn{3}{|c|}{$MD = .10$} & $IR$ & $OR$ & $b_{p,+}$\\
        \hline
        $f_+=.5$ & $f_{p,+}=.41$ & $f_{u,+}=.51$ & \lightgreen{$.804$} & \bcolor{red}{$.668$} & \bcolor{red}{$.18$}\\
        \hline
        $f_+=.1$ & $f_{p,+}=.01$ & $f_{u,+}=.11$ & \bcolor{red}{$.091$} & \bcolor{red}{$.082$} & \bcolor{red}{$.9$}\\
        \hline
    \end{tabular}
    \caption{Example data used by \cite{oswald2016measuring}.}
    \label{tab:oswald_IR_OR_b}
\end{table} 

\section{Intersectionality and Multi-class Problems}
\label{sec:intersect}
In this section, we extend our approach to consider multi-valued labels, as well as to be intersectional, \ie to consider multiple sensitive groups and their intersections in the analysis. This approach is critical when assessing fairness in real world applications~\cite{hort2024biasmitigationsurvey}. %

\begin{example}
    The need to consider non-binary labels comes very naturally in practice; as a paradigmatic example, we will consider the COMPAS dataset \cite{ds_compas} containing records for US criminal offenders and a score of their likelihood to reoffend (recidivism). The scores are given using three labels (low, medium, or high), so it would be ideal to capture this with our bias measure as well. The same goes for having multiple (potentially non-binary) sensitive attributes, in this case we consider two binary sensitive attributes gender and race, taking values in \{men, women\} and \{white, non-white\} respectively. We denote men with $m$, women with $w$, white tuples with $c$ (``caucasian'') and non-white ones with $o$ (for ``others'').  The data is summarized in Table~\ref{tab:compas_ini_values}.
\end{example}

One of the main benefits of our measure is that extending it to non-binary labels and multiple sensitive attributes is 
natural, as we can see in Definition \ref{def:UB_multiattr}.

\begin{definition}[\MeasureName , multi attributes, general label]\label{def:UB_multiattr}
    Given a dataset $T$ with sensitive attributes $\textbf{S}$
    we say that the Universal Bias of group $\textbf{s}\in Dom(\textbf{S})$ w.r.t. label $y\in\{y_1,\dots ,y_k\}$ is 
    \[ 
        b_{\textbf{s},y} = 1-\frac{f_{\textbf{s},y}}{f_y}~.    
    \]
\end{definition}

Note how Definition \ref{def:UB_multiattr} naturally leads to the generalized version of an unbiased dataset, given in Definition \ref{def:unbiased_ds_multiattr}.

\begin{definition}[Unbiased dataset, multi attributes, general  label]\label{def:unbiased_ds_multiattr}
    In the setting of Definition \ref{def:UB_multiattr},
    we say that $T$ is unbiased w.r.t. $\textbf{S}$ and label $y\in\{y_1,\dots ,y_k\}$ if for every $\textbf{s}\in Dom(\textbf{S})$
    \[
        f_{\textbf{s},y} = f_y 
    \]
\end{definition}

\begin{remark}
Recall our notational convention explained in Section \ref{sec:prelim} on the use of the value $\epsilon$ for a sensitive attribute.
\end{remark}

\begin{example}\label{ex:adult_2attr}
In our COMPAS example, we have eight possible protected groups: four that only consider one attribute $\{m,w,c,o\}$, and four binary $\{mc,mo,wc,wo\}$. For each group and label, the data is unbiased if the frequency of the group having that label is equal to the frequency of the entire population having that label. %
\end{example}

Note that as before having $b_{\textbf{s},y}=0$ for all groups $\textbf{s}$ and label values is equivalent to the unbiased condition stated in Definition \ref{def:unbiased_ds_multiattr}. An important feature of UB is that we can analyze the bias of each group (and label) separately.
To illustrate the usefulness of our techniques, we will use \MeasureName\ to analyze the biases in the COMPAS dataset. 

\begin{table}[t]
    \centering
    \begin{tabular}{|c|c|c|c|c|}
          \hline
        {}  & {label} & o & c & Total\\
          \hline
        {} & L & 19489 & 12202 & 31691 \\
        {m} & M & 7143 & 2862 & 10005 \\
        {} & H & 4510 & 1273 & 5783 \\
        {} & Tot. & mo: 31142 & mc: 16337 & m: 47479\\
        \hline
        {} & L & 5637 & 4159 & 9796 \\
        {w} & M & 1589 & 894 & 2483 \\
        {} & H & 665 & 375 & 1040 \\
        {} & Tot. & wo: 7891 & wc: 5428 & w: 13319 \\
        \hline
        {} & L & 25126 & 16361 & 41487 \\
        {Total} & M & 8732 & 3756 & 12488 \\
        {} & H & 5175 & 1648 & 6823 \\
        {} & Tot. & o: 39033 & c: 21765 & n: 60798 \\
        \hline
    \end{tabular}
    \caption{Summary statistics (values) for initial version of the COMPAS dataset with ternary label.}
    \label{tab:compas_ini_values}
\end{table}

\begin{table}[t]
    \centering
    \begin{tabular}{|c|c|c|c|c|}
          \hline
        {}  & {label} & o & c & Total\\
          \hline
        {} & L & 0.083 & -0.095 & 0.022 \\
        {m} & M & -0.117 & 0.147 & -0.026 \\
        {} & H & -0.290 & 0.306 & -0.085 \\
        \hline
        {} & L & -0.047 & -0.123 & -0.078 \\
        {w} & M & 0.020 & 0.198 & 0.092 \\
        {} & H & 0.249 & 0.384 & 0.304 \\
        \hline
        {} & L & 0.057 & -0.102 & 0 \\
        {Total} & M & -0.089 & 0.160 & 0 \\
        {} & H & -0.181 & 0.325 & 0 \\
        \hline
    \end{tabular}
    \caption{Summary statistics (biases) for initial version of the COMPAS dataset with ternary label.}
    \label{tab:compas_ini_biases}%
\end{table}

\begin{example}\label{ex:compas_discr}
The values and biases of the COMPAS dataset are shown in Table \ref{tab:compas_ini_values} and Table \ref{tab:compas_ini_biases} respectively.

This dataset is known to have multiple disparities in the treatment of different sensitive groups. This is the case for men and women, where the ratio of women with a low risk score is substantially greater than the ratio of men (\ie $f_{w,L}>f_{m,L}$). It is also the case that $f_{w,M}<f_{m,M}$ and $f_{w,H}<f_{m,H}$. This case can be numerically quantified with \MeasureName, 
$b_{m,L} = .022>0$ (bias against men) and 
$b_{w,L} = -.078<0$ (bias in favor of women). There is also a noticeable difference when the race attribute is considered: the magnitude of the bias of non-white people with a high score is half that of the one among white people ($b_{o,H}=-.181$ and $b_{c,H}=.325$). Note how our measure naturally captures what happens in terms of the frequencies: the rate of people with a high score among the non-white population is double the rate among white people (using the values in Table \ref{tab:compas_ini_values}, $f_{c,H}=.076$ and $f_{o,H}=.133$).
We observe a difference in terms of the gender attribute as well, namely $b_{w,H}=.304$ and $b_{m,H}=-.085$ ($f_{w,H}=.078$ and $f_{m,H}=.122$).
\end{example}

There is a consensus that the difference in the race attribute constitutes actual discrimination \cite{hort2024biasmitigationsurvey}. This may not be the case for the difference found for the gender attribute (not reported for this dataset). This case could either be a true social phenomenon that is not to be corrected 
or discrimination. Within our framework, this can be done using external sources to, for example, determine that in a certain population $b_{w,H}$ may not be zero.
In this example, sociologists may decide a value of $0.304$ is a reasonable societal norm and therefore this dataset is not biased for this class.
This is the type of decision an expert in the domain should make. Our mathematical methods are designed as a tool to help experts make these decisions. The mitigation algorithms we define next can be used to create a dataset with zero bias for a specific group or with a bias that is {\em a priori} set by experts. We will extend our techniques to be able to model these alternatives in Section \ref{sec:extensions}. 
\section{General data bias mitigation}\label{sec:mitigation}

We introduce our general mitigation algorithm, prove its correctness and use it to mitigate the COMPAS dataset. %

Let $T$ be a dataset with sensitive attributes $\textbf{S}$ and label $Y$ with values $y\in\{y_1,\dots ,y_k\}$. In the context of bias mitigation, we want to add or delete tuples to reduce the data bias as much as possible. We note that one can always consider tuple deletions as a preprocessing step, \ie first delete some of the tuples and then measure bias and determine the new tuples that should be added for mitigation. Thus, we will only consider tuple additions in this context. We denote the number of tuples to be added from group \textbf{s} with label value $y$ by $\Delta \textbf{s}y$. The key idea for the algorithm comes from using Definition \ref{def:unbiased_ds_multiattr}, since we want to determine $\Delta \textbf{s}y$ such that this definition holds, we can write this as
\begin{equation}\label{eq:mitig}
    f_{\textbf{s},y} = \frac{\textbf{s}y+\Delta \textbf{s}y}{\textbf{s}+\Delta \textbf{s}} = f_y \Leftrightarrow \frac{\textbf{s}y+\Delta\textbf{s}y}{f_y} = \textbf{s}+\Delta \textbf{s}
\end{equation}
This condition must hold for every group \textbf{s} and label value $y$, so we have a system of equations, where the unknowns are $\Delta \textbf{s}y$. 
The label frequency preservation that we introduce later is a direct consequence of the equation on the left. Then since the equality on the right in Equation (\ref{eq:mitig}) holds for every $y$, we can write
\begin{equation}\label{eq:mitig2}
    \frac{\textbf{s}y+\Delta\textbf{s}y}{f_y} = \frac{\textbf{s}y_i+\Delta\textbf{s}y_i}{f_{y_i}}
\end{equation}
for every $y\in\{y_1,\dots,y_k\}$. That is we equate all equations with the one corresponding to $y_i$. Solving for the unknown we are looking for we get, $\Delta\textbf{s}y=-\textbf{s}y + \frac{f_y}{f_{y_{i}}}(\textbf{s}y_i+\Delta \textbf{s}y_i)$. The full solution of the system is presented in Theorem \ref{thm:general_mitigation}.\footnote{All proofs are included in the Appendix (Section \ref{sec:proofs}).}

\begin{theorem}\label{thm:general_mitigation}
    Given a dataset $T$ with sensitive attributes $\textbf{S}$ and label $Y$ with values $y\in\{y_1,\dots ,y_k\}$ and index $i = \arg\max_j \frac{\textbf{s}y_j}{y_j}$.
    The general solution for the system given by Equation (\ref{eq:mitig2}) for every $y$ is as follows
    \[
        \Delta \textbf{s}y = -\textbf{s}y + \Big\lfloor\frac{f_y}{f_{y_i}}(\textbf{s}y_i+\Delta \textbf{s}y_i)\Big\rfloor %
    \]
    for every group \textbf{s} and label value $y$, for $\Delta sy_i\geq 0$. For a given value of $\Delta sy_i$ (free variable), this solution determines the number of tuples that need to be added to each sensitive group to produce a mitigated dataset $T_m$ that is unbiased. Specifically, $T_m$~will contain $\Delta\textbf{s}y$ new tuples from group $\textbf{s}$ with label $y$. Note that we do not consider tuples with missing values, \ie all the added tuples need to have a value for each sensitive attribute.  %
\end{theorem}

\begin{remark}
    The solution from Theorem \ref{thm:general_mitigation} can be written as 
    \[
        \textbf{s}y^{\text{mit}}:= \Delta \textbf{s}y+\textbf{s}y = \Big\lfloor \frac{f_y}{f_{y_i}}(\textbf{s}y_i+\Delta \textbf{s}y_i)\Big\rfloor = \Big\lfloor \frac{f_y}{f_{y_i}} \textbf{s}y_i^{\text{mit}}\Big\rfloor
    \]
     where $\textbf{s}y^{\text{mit}}$ is the total number of tuples belonging to group \textbf{s} with label $y$ in the mitigated dataset. Note that since $sy\in\mathbb{N}$ then $\lfloor -\textbf{s}y\rfloor = -\textbf{s}y$. 
\end{remark}

\begin{example}
    For COMPAS, one such solution is shown in Table \ref{tab:compas_mitigated}.
Using our formulas we can check that this new version is unbiased for all sensitive groups.  
Note how in order to mitigate the gender biases, we needed to add more men than women ($\Delta m=10592$ and $\Delta w=1037$). However, when disaggregating the groups, we see that the majority of added men have a low risk score ($\Delta mL=7935$) and all added women have medium or high scores ($\Delta wM=466$ and $\Delta wH=571$). When also considering race, we see that all but $1$ added white men have medium or high risk scores ($\Delta mcM=811$ and $\Delta mcH=734$) and that the majority of added non-white males have a low risk score ($\Delta moL=7934$). This showcases the importance of analyzing tuple additions for the different intersectional groups. 
\end{example}

\begin{table}[]
    \centering
    \begin{tabular}{|c|c|c|c|c|}
          \hline
          & {label} & o & c & Total\\
          \hline
        {} & L & 27422   & 12202  & 39624  \\
        {m} & M & 8254   & 3672 &  11926 \\
        {} & H & 4510   & 2006 &  6516 \\
        {} & Tot. & mo: 40186 & mc: 17880 & m: 58066\\
        \hline
        {} & L &  5637  &  4159 & 9796 \\
        {w} & M &  1697  & 1251 & 2947  \\
        {} & H &  927  & 683 & 1610 \\
        {} & Tot. & wo: 8260  & wc: 6093  & w: 14353\\
        \hline
        {} & L & 33059 & 16361 & 49420 \\
        {Total} & M & 9950 & 4923 & 14873 \\
        {} & H & 5437 & 2689 &  8126 \\
        {} & Tot. & o: 48446 & c: 23973 & n: 72419\\
        \hline
    \end{tabular}
    \caption{Summary statistics (values) for a mitigated version of the COMPAS dataset without deletions. Note that this table is unbiased, for all eight groups, 
    \ie $f_{\textbf{s},L}=f_L=.682$, $f_{\textbf{s},M}=f_M=.205$ and $f_{\textbf{s},H}=f_H=.112$.}
    \label{tab:compas_mitigated}
\end{table}

A solution for the equations in Theorem \ref{thm:general_mitigation} determines a number of tuples of each group to make $T$ unbiased. But what is the minimal solution, \ie a solution with the least number of additions? Corollary \ref{cor:mitig_minimality} gives the answer.

\begin{corollary}\label{cor:mitig_minimality}
    In the same setting as Theorem \ref{thm:general_mitigation}, the least number of tuple additions for every group \textbf{s} and label value $y$ for making $T$ unbiased is given by
    \[
        \Delta \textbf{s}y + \textbf{s}y = \Big\lfloor y\frac{\textbf{s}y_i}{y_i}\Big\rfloor \Leftrightarrow \textbf{s}^{\text{mit}} = \frac{\textbf{s}y_i}{y_i}n+ C \text{, }0\leq C<k
    \]
    where $\textbf{s}^{\text{mit}}$ is the total number of tuples belonging to group \textbf{s} in the mitigated dataset and $n$ is the total number of tuples in the initial version. Note how $C$ is a small constant bounded by the number of classes $k$.
\end{corollary}

\begin{example}
    In fact, Table \ref{tab:compas_mitigated}, is a minimal solution. When mitigating this dataset with our minimal solution, we have for $\textbf{s}=mo$, $y_i=H$ and for the rest of the groups ($mc$, $wo$ and $wc$) $y_i=L$. Note how ${wo}^{\text{mit}} = \frac{{wo}L}{L}n=\lfloor (5637/41487)\cdot 60798\rfloor=8260$.
\end{example}

\subsection{Label frequency preservation property}

There are different statistical properties that a data scientist could want to preserve when altering the data in the mitigation process. A intuitive property
that is easily motivated by the COMPAS example is the preservation of the label frequencies, hence this is the one we guarantee with our algorithm and prove in Theorem \ref{thm:lfpp}. Note that this property provides a mathematical guarantee for the mitigated data to be used in the same context as the original data. Most ML algorithms and statistical methods rely on assumptions about the input distribution, so it is of key importance to preserve the representativeness of the data as much as possible for these downstream tasks. 

\begin{theorem}(Preservation of label frequencies)\label{thm:lfpp}
    The mitigation algorithm given by Theorem \ref{thm:general_mitigation} preserves the label frequencies among the total population ($f_{y_j}$) from the original dataset.
\end{theorem}

\begin{example}
    Recall the values (Table \ref{tab:compas_ini_values}) for the COMPAS dataset. We have $f_L = 41487/60798 \approx .682$,, $f_M\approx.205$ and $f_H\approx.112$. Note that in the mitigated version (Table \ref{tab:compas_mitigated}) these frequencies are preserved (\eg $f_L = 49420/72419 \approx .682$).
\end{example}
Since in the original dataset (Table \ref{tab:compas_ini_values}), $68\%$ of the population has low recidivism risk score and only $11\%$ has high risk, we would ideally want to preserve this feature in the mitigated version of the data (since this can be a crucial feature of it for the downstream task). As a consequence, we feel that if for example after mitigation we get $f_H\approx .25$ and $f_L\approx .35$, we would be significantly changing a key feature of the data.
In general, having the option to keep the original frequencies of the label values is a desirable feature of a mitigation algorithm.

\section{Practical Generalizations}
\label{sec:extensions}

Now we extend our framework to permit the unbiased condition to be non-zero. Incorporating costs (or weights) on tuples of different groups into the mitigation algorithm is deferred to the appendix (Section \ref{sec:costs}).

As mentioned in Section \ref{sec:intersect}, while it is common to define a zero bias condition, this is not always the desired unbiased goal in practice.
The reason is simple: a domain expert may consider that a given bias, for example the difference between women and men in Example \ref{ex:compas_discr}, represents a social behavior that does not constitute discrimination. As a consequence, a mitigation algorithm should take this fact into consideration. We therefore generalize our unbiased conditions to allow a data scientist to specify a number $K_{\textbf{s},y}>0$ for each group \textbf{s} and label value $y$, which is the desired ratio of the success rate of the protected group to the success rate of the population:
\[
    \frac{f_{\textbf{s},y}}{f_{y}}=K_{\textbf{s},y}
\]
To recover the original setting (where bias of 0 is the goal), we can set $K_{\textbf{s},y}=1$. For the selected values of $K_{\textbf{s},y}$, the following identity holds $\sum K_{\textbf{s},y}f_{y}=1$.  %
\begin{example}
    In Table \ref{tab:compas_mitigated_Ks}, we show the summary statistics of a mitigated version of the COMPAS dataset with gender and race protected attributes with values $\textbf{s}=\langle g,r\rangle$ and the following $K$ values: $K_{gr,y}=\frac{f_{g,y}}{f_y}$, chosen such that 
    \[
      f_{gr,y}=\frac{gry+\Delta gry}{gr + \Delta gr} = f_{g,y}
    \]
    meaning that after mitigation is applied the proportion of any gender group is the same as prior to mitigation. Note that if $g=\epsilon$ (\ie we only consider race), then $K_{gr,y}=f_y/f_y=1$ as before.
\end{example}

\begin{table}[]
    \centering
    \begin{tabular}{|c|c|c|c|c|}
          \hline
          & {label} & o & c & Total\\
          \hline
        {} & L &  24715  & 12204  & 36919 \\
        {m} & M &  7803  & 3853 &  11656 \\
        {} & H &  4510  & 2227 &  6737 \\
        {} & Tot. & mo: 37028  & mc: 18284 & m: 55312\\
        \hline
        {} & L &  6274  & 4161 & 10435 \\
        {w} & M &  1590  & 1055 &  2645 \\
        {} & H &  668  & 443 & 1111 \\
        {} & Tot. & wo: 8532 & wc: 5659 & w: 14191\\
        \hline
        {} & L & 30989 & 16365 & 47354 \\
        {Total} & M & 9393 & 4908 & 14301 \\
        {} & H & 5178 & 2670 & 7848 \\
        {} & Tot. & o: 45560 & c: 23943 & n: 69503\\
        \hline
    \end{tabular}
    \caption{Summary statistics (values) of a mitigated version of the COMPAS dataset with ternary labels using constants $K_{\textbf{s},y}\neq 1$.}
    \label{tab:compas_mitigated_Ks}
\end{table}

\section{ML based evaluation}
\label{sec:mleval}
In this section, we train ML models on the mitigated and biased versions of the datasets to see how our data bias mitigation strategies affect ML performance.

\subsection{Methodology}

In order to simulate the setting in which we have a biased dataset $T$ and we want to collect external data (\eg from an open data lake) to mitigate a bias $T$ has, we partition $T$ into two sets uniformly at random: an ``initial sample'' of size $x_0$ and the remaining portion that we consider to be the external available data. Similar as Salimi et al. \cite{salimi2020database}, we then generate mitigated versions of the dataset using different mitigation strategies if possible. We distinguish between two types of mitigation strategies: a strategy that preserves the initial label frequencies of the data and one that does not. 
Then, for each mitigated dataset, we generate a uniform sample of the same initial size.
We train a set of ML models on both samples and evaluate their accuracy, precision and recall. Training is done using a random $80\%$ portion of the dataset for training and the remaining $20\%$ for evaluation.

\subsection{Datasets \& Models used}

We use three widely known datasets for our experiments: the Adult dataset \cite{misc_adult_2} introduced in Section \ref{sec:intro}, the Default dataset \cite{misc_default_of_credit_card_clients_350} containing information about default payments of credit card clients in Taiwan and the COMPAS dataset \cite{ds_compas} introduced in Section \ref{sec:intersect}. We apply the standard preprocessing techniques: we normalize numeric features and use one-hot encoding for categorical features.\footnote{The source code is available at \url{https://github.com/bscarone/data-bias-pub}.} 
In terms of ML models, we use Random Forest, Gradient Boost Decision Tress (GBDT), Extra Trees, Ada Boost, Multilayer Perceptron (MLP) and Logistic Regression. We use the open source ML library Scikit-learn\footnote{Scikit-learn documentation, \url{https://scikit-learn.org/stable/}. Accessed: 2024-12-10} to implement our learning algorithms.

\subsection{Experimental results}
We report results for one (different) model for each dataset, as the other experiments exhibit similar results. We segment the charts into two regions, the left containing the results for the unbiased versions of the dataset and the right the biased samples. We use the label ``uX\%" for uniform samples of the data of size $X\%$, the label ``nX\%" for a mitigated sample that does not preserve the initial label frequencies of the data and ``pX\%" for the samples that do.

\begin{figure}[t]
    \centering
    \includegraphics[scale=0.35]{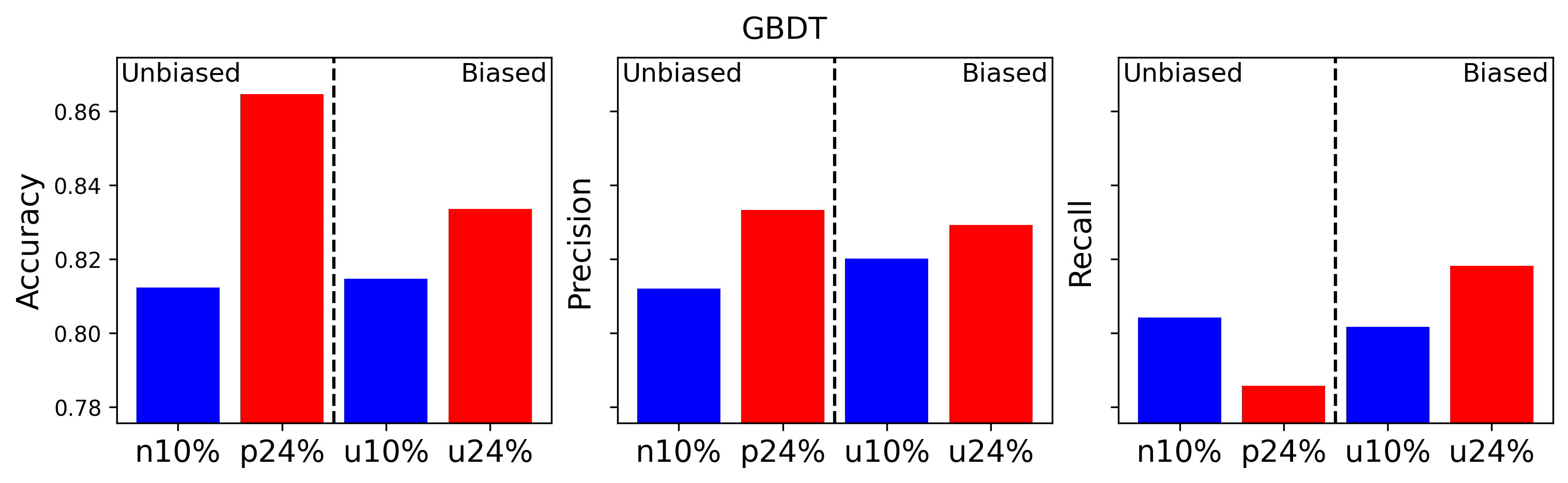}
    \caption{Adult dataset - GBDT ML model.}
    \label{fig:adult_mleval}
\end{figure}

\begin{figure}[t]
    \centering
    \includegraphics[scale=0.35]{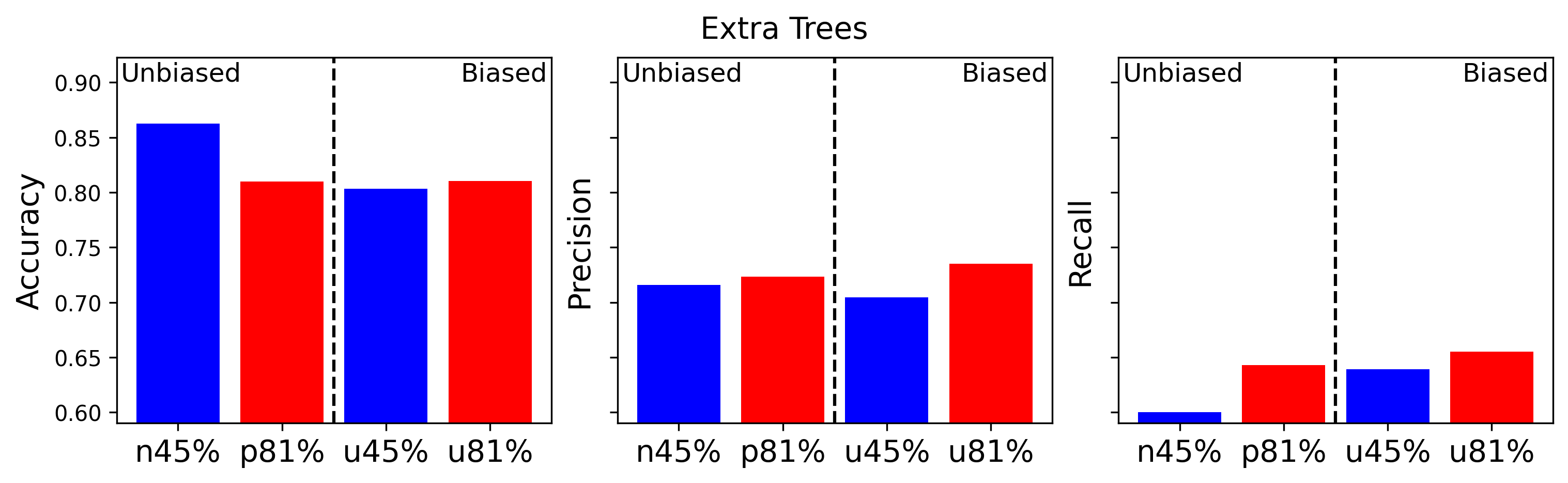}
    \caption{Default dataset - Extra Trees ML model.}
    \label{fig:default_mleval}
\end{figure}

\begin{figure}[t]
    \centering
    \includegraphics[scale=0.35]{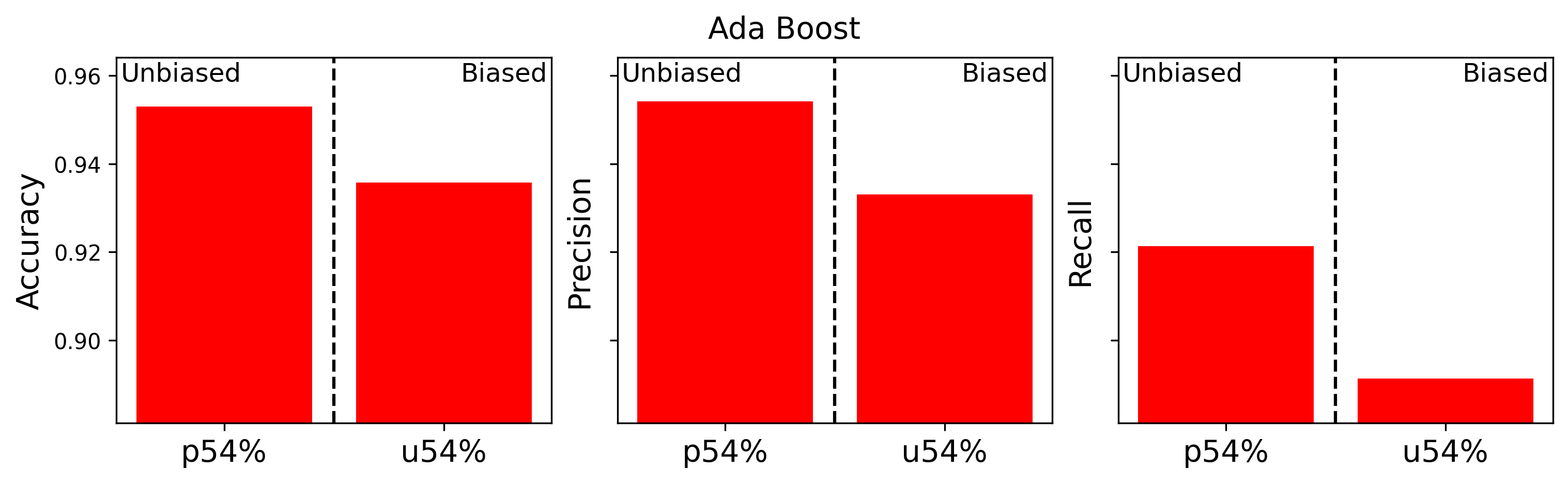}
    \caption{COMPAS dataset - Ada Boost ML model.}
    \label{fig:compas_mleval}
\end{figure}

For the Adult dataset (Fig.~\ref{fig:adult_mleval}), the accuracy when using a mitigated dataset that preserves the label frequencies is significantly higher. Precision increases slightly and Recall decreases slightly. For the Default dataset (Fig.~\ref{fig:default_mleval}), similar accuracy is observed for all measures, only the accuracy for the mitigated version that does not preserve label frequencies is a bit higher. %
Surprisingly good results are observed for the COMPAS dataset (Fig.~\ref{fig:compas_mleval}), where we consider ternary labels (instead of making them binary as is usually done in previous work \cite{hort2024biasmitigationsurvey}), 
using a mitigation that preserves the initial label frequencies of the data. In this case, the performance of the models is always consistently better than that of a uniform sample for all measures.

For the binary label case, when comparing our results to those reported by Salimi et al. \cite{salimi2020database} we can eliminate \textbf{all} the data bias without significant performance losses. What is more surprising is that for the case of ternary labels, we even improve the performance of all ML models when mitigating the bias. 
Note that although Salimi et al. introduce a bias mitigation technique, they measure bias using a slightly modified conditional version of the OR (Odds Ratio), one of the existing measures introduced in Section \ref{sec:prelim}. Only an intuitive justification for this choice is given and their repair algorithm does not provide direct control over bias. 
We have argued why UB is a better measure than the OR in the binary setting in Section \ref{sec:motivation}.  Importantly, there is no version of OR for the general setting (multiple non-binary sensitive attributes and label values), while our UB directly translates to this setting (Section \ref{sec:intersect}). We note that we can completely eliminate the bias in our dataset, \ie we reach $UB(T,\textbf{s},y)=0$ for every group $\textbf{s}$ and label value $y$, which in the binary case corresponds to $OR=1$.
\section{Conclusions and Future Work}
\label{sec:conclusion}

We present \MeasureName, the first interpretable and intersectional multi-label bias measure that can be computed directly and efficiently from a given dataset. To exemplify this we show how \MeasureName{} solves issues in the anti employment discrimination rules currently being used by the US Department of Labor. 
The same ideas used to derive UB can serve as the basis for an explainable and mathematically guaranteed bias mitigation algorithm that can be computed for any dataset. Our algorithm preserves the label frequencies, a mathematical guarantee for the mitigated data to be used in the same context as the original data.

We evaluate our techniques on widely used real datasets and show that our mitigation strategies can be used in practice to produce training data that yield better performing ML models. 

In the future, we want to explore the relation between the bias of each group and the ML performance for the group. We plan to explore extensions of table union search~\cite{DBLP:conf/sigmod/Fan00M23} that incorporate the number of tuples from a group or with a particular label in the retrieved tables as part of the ranking criteria. This will allow us to implement scalable versions of our mitigation strategy over real data lakes.
With our techniques, one can also obtain mitigated datasets that do not preserve the label frequencies. 
We would like to explore contexts where a solution without this property would be better than not mitigating the bias at all. For example, we can incorporate the differences in the distributions as another optimization criterion.
We also want to consider the context in which we have missing values in our data, where imputation techniques may be useful. 

\bibliography{bibliography}

\clearpage 
\appendix 

\section{Omitted Proofs}\label{sec:proofs}

\begin{theorem}
    Given a dataset $T$ with sensitive attributes $\textbf{S}$ and label $Y$ with values $y\in\{y_1,\dots ,y_k\}$ and index $i = \arg\max_j \frac{\textbf{s}y_j}{y_j}$.
    The general solution for the system given by Equation (\ref{eq:mitig2}) for every $y$ is as follows
    \[
        \Delta \textbf{s}y = -\textbf{s}y + \Big\lfloor\frac{f_y}{f_{y_i}}(\textbf{s}y_i+\Delta \textbf{s}y_i)\Big\rfloor %
    \]
    for every group \textbf{s} and label value $y$, for $\Delta sy_i\geq 0$. For a given value of $\Delta sy_i$ (free variable), this solution determines the number of tuples that need to be added to each sensitive group to produce a mitigated dataset $T_m$ that is unbiased. Specifically, $T_m$~will contain $\Delta\textbf{s}y$ new tuples from group $\textbf{s}$ with label $y$. Note that we do not consider tuples with missing values, \ie all the added tuples need to have a value for each sensitive attribute.  %
\end{theorem}
\begin{proof}%
    In the system of equations we have $k-1$ equations (for $y=y_i$ we have $0=0$) with $k$ unknowns, where $\Delta \textbf{s}y_i$ can be taken as the free variable. The solution, obtained directly from Equation (\ref{eq:mitig2}), can be written as $\Delta \textbf{s}y = A + B\Delta \textbf{s}y_i$ where $A=-\textbf{s}y+\frac{y}{y_i}\textbf{s}y_i$ and $B=\frac{f_y}{f_{y_i}}=\frac{y}{y_i}$.\\

    Since we are adding tuples, it must be the case that all $\Delta \textbf{s}y$ are non-negative integers, \ie 1) $\Delta \textbf{s}y\geq 0$ and 2) $\Delta \textbf{s}y\in\mathbb{N}$. For 1) we need to prove that for any $y$, $A\geq 0\Leftrightarrow \frac{\textbf{s}y_i}{y_i}\geq \frac{\textbf{s}y}{y}$. That is, it suffices to take $i = \arg\max_j \frac{\textbf{s}y_j}{y_j}$ as is done in the Theorem. For 2), we use the floor operator, noting that we are allowed to do so because the solution is non-negative. Then, by taking any non-negative integral value of $\Delta \textbf{s}y_i$, we can substitute it in the other equations to determine the rest of the $\Delta \textbf{s}y_j$. 
\end{proof}

\begin{corollary}
    In the same setting as Theorem \ref{thm:general_mitigation}, the least number of tuple additions for every group \textbf{s} and label value $y$ for making $T$ unbiased is given by
    \[
        \Delta \textbf{s}y + \textbf{s}y = \Big\lfloor y\frac{\textbf{s}y_i}{y_i}\Big\rfloor \Leftrightarrow \textbf{s}^{\text{mit}} = \frac{\textbf{s}y_i}{y_i}n+ C \text{, }0\leq C<k
    \]
    where $\textbf{s}^{\text{mit}}$ is the total number of tuples belonging to group \textbf{s} in the mitigated dataset and $n$ is the total number of tuples in the initial version. Note how $C$ is a small constant bounded by the number of classes $k$.
\end{corollary}
\begin{proof}%
Analogously as in Theorem \ref{thm:general_mitigation} we have $\Delta \textbf{s}y = A + B\Delta \textbf{s}y_i$ where $A=-\textbf{s}y+\frac{y}{y_i}\textbf{s}y_i$ and $B=\frac{y}{y_i}$. Since by the choice of $i$ we have $A\geq 0$ for every label value $y$, then we can also set $\Delta \textbf{s}y_i =0$ in every equation
without the risk of $\Delta \textbf{s}y$ becoming negative, which produces the minimal solution.
A nice consequence of this minimal formulation is that when one sums both sides of $\Delta \textbf{s}y +\textbf{s}y = \Big\lfloor y\frac{\textbf{s}y_i}{y_i}\Big\rfloor$ for all $y$, one gets $\textbf{s}^{\text{mit}} = \frac{\textbf{s}y_i}{y_i}n + C$ with $0\leq C<k$, \ie a simple expression\footnote{Normally the number of classes $k$ is around 5, while the number of tuples is typically bigger than $5000$ meaning that this is a good approximation.} for the total number of tuples of group \textbf{s} in the mitigated version in terms of quantities of the old table.
\end{proof}

\begin{remark}
    We can generalize Theorem \ref{thm:general_mitigation} and Corollary \ref{cor:mitig_minimality} for the context where we have costs, a budget and the $K_{\textbf{s},y}$ constants.
\end{remark}

\begin{theorem}(Preservation of label frequencies)
    The mitigation algorithm given by Theorem \ref{thm:general_mitigation} preserves the label frequencies among the total population ($f_{y_j}$) from the original dataset.
\end{theorem}
\begin{proof}%
    This property is a direct consequence of the equations of the system: namely, we are adding $\Delta \textbf{s}y_j$ tuples such that for every $1\leq j\leq k$, $f_{\textbf{s},y_j} =\frac{\textbf{s}y_j+\Delta \textbf{s}y_j}{\textbf{s}+\Delta \textbf{s}}=f_{y_j}$, where $f_{y_j}=y_j/n$ is the ratio of tuples with label value $y_j$ in the original dataset. 
\end{proof}

\section{Practical Generalizations - Costs and Budget}
\label{sec:costs}
An initial way to add costs inspired by the data discovery literature is to assume that searching for a tuple of type $\textbf{s}y$ has cost $c_{\textbf{s}y}$. So, $\Delta \textbf{s}y\cdot c_{\textbf{s}y}$ is the total cost of adding such tuples. With this extension, one can consider a budget $B$ and restrict the search for mitigation solutions with the condition $\sum_{\textbf{s},y} \Delta \textbf{s}y\cdot c_{\textbf{s}y}\leq B$.

\begin{example}
With our techniques it is easy to solve the examples from Figure \ref{fig:adult_2varspols} analytically. In this example, all tuples have the same costs and we have the following budgets and restrictions over the total number of tuples. For the case of Figure \ref{fig:adult_2varspols_sub2}, $a_{p+}\leq 3000$ and $n^{\text{mit}}=p+u+a_{p+}-d_{u-}\geq 30,000$. Then, to compute the \MeasureName\ analytically in terms of $a_{p+}$ and $d_{u-}$ we do the following: compute $f_{p,+}=\frac{p^++a_{p+}}{p+a_{p+}}$, $f_{u,+}=\frac{u^+-d_{u+}}{u-d_{u+}}$ and $f_{+}=\frac{p^++u^++a_{p+}-d_{u+}}{n+a_{p+}-d_{u+}}$ based on the number of tuples of the table and the number of added and deleted tuples, and then one can directly compute $b_{p,+}=1-f_{p,+}/f_+$. This gives a complete analytical characterization of all the possible mitigated tables and their biases using the strategy of adding positive protected tuples and deleting positive unprotected ones. We note that this strategy does not preserve label frequencies. If one wants to preserve this property, one can use our mitigation algorithm.
\end{example}

\begin{example}
Now we will consider the COMPAS dataset under budget and cost restrictions. After computing the mitigated version of the data (Table \ref{tab:compas_mitigated}), we can determine what tuples needed to be added (comparing against Table \ref{tab:compas_ini_values}): $\Delta w=872= 14191 - 133119$, $\Delta mL=5228$, $\Delta mM = 1651$ and $\Delta mH=954$. Suppose we have a Budget of $B=7500$.

If all costs are one, $c_{\textbf{s}y}=1$, and we want to mitigate the bias against women, there are $7500-872 = 6628$ tuples remaining to mitigate the bias against men. This budget is enough for example to mitigate the bias against men with a low and high labels ($mL$ and $mH$). After this, the remaining budget is $6628 - 5228 - 954 = 1205$. For mitigating the bias against men with a medium score ($mM$) we would need $1651$ tuples, but we are only able to add $1205$ as a partial mitigation in this case.

Now suppose that the cost of any women tuple is $2$ and all the other tuples remain with a cost of $1$. Mitigating the bias against women leaves a budget of $7500-2\cdot 872 = 5756$. Then we can decide to use it to mitigate the bias of $mM$ and $mH$, which leaves us with $5756-955-1651=3150$. We would need $2077$ additional tuples to mitigate the bias of $mL$. That is, one can use our algorithm to first compute the mitigated version and then using the allowed budget compare different scenarios for partial mitigation. 
\end{example}

\end{document}